\documentclass[conference,a4paper]{IEEEtran}
\IEEEoverridecommandlockouts

\usepackage{cite}
\usepackage{amsmath,amssymb,amsfonts}
\usepackage{algorithmic}
\usepackage{graphicx}
\usepackage{textcomp}
\usepackage{xcolor}
\usepackage{hyperref}
\def\BibTeX{{\rm B\kern-.05em{\sc i\kern-.025em b}\kern-.08em
    T\kern-.1667em\lower.7ex\hbox{E}\kern-.125emX}}

\usepackage{eso-pic}
\AddToShipoutPictureFG*{%
  \AtPageUpperLeft{%
    \put(\LenToUnit{\dimexpr\paperwidth-1.6cm\relax},\LenToUnit{-1.2cm}){%
      \makebox(0,0)[r]{\footnotesize 2026 IEEE Region 10 Conference (TENCON)}%
    }%
  }%
}

\IEEEpubid{%
  \parbox{\textwidth}{%
    \vspace{5mm}
    \fontsize{8.5}{10}\selectfont\raggedright
    This paper is accepted by the 2026 IEEE Region 10 Conference (TENCON).
    Copyright \copyright{}2026 IEEE. Personal use of this material is permitted.
    However, permission to use this material for any other purposes must be
    obtained from IEEE by sending an email to pubs-permissions@ieee.org.%
  }%
}

\definecolor{deepblue}{RGB}{0, 70, 140}
\definecolor{deepgreen}{RGB}{0, 120, 60}

\title{ConvRoute-ViT: Convolutional Token Routing for Efficient Vision Transformers}

\begin{document}
\author{
\IEEEauthorblockN{Manish Das}
\IEEEauthorblockA{\textit{IIIT Allahabad} \\
Prayagraj, India \\
pmm2023001@iiita.ac.in}
\and
\IEEEauthorblockN{Shiv Ram Dubey}
\IEEEauthorblockA{\textit{IIIT Allahabad} \\
Prayagraj, India \\
srdubey@iiita.ac.in}
\and
\IEEEauthorblockN{Satish Kumar Singh}
\IEEEauthorblockA{\textit{IIIT Allahabad} \\
Prayagraj, India \\
sk.singh@iiita.ac.in}
\thanks{We gratefully acknowledge the Indian Institute of Information Technology Allahabad for funding and providing necessary support to pursue this research work.}
}

\maketitle

\begin{abstract}
Vision Transformers (ViTs) achieve strong image recognition performance, but their self-attention increases quadratically with the number of input tokens and makes them computationally expensive. The standard implementation of ViT models treats all patch tokens with equal importance across every layer of the model. This uniform treatment introduces redundant computation and may reduce the model's focus on the most discriminative object regions. In this paper, we propose ConvRoute-ViT, a convolutional token routing mechanism for efficient vision transformers that performs token reduction through convolutional routing. Our model preserves the most relevant patch tokens according to class-token attention at selected layers, while the remaining tokens are spatially reorganized and compressed into a compact set of convolutional summary tokens. This design reduces the sequence length without completely removing the contextual information of the less informative tokens. By applying the routing process hierarchically, the proposed framework constructs compact and discriminative representations for deeper layers. Experiments on CIFAR-100 and TinyImageNet show that ConvRoute-ViT improves computational efficiency while maintaining classification accuracy compared with the base ViT.
\end{abstract}


\section{Introduction}
Vision Transformers (ViTs) have emerged as an effective alternative to convolutional neural networks for image recognition by modeling global relationships across diverse computer vision tasks \cite{b1}. The complexity of self-attention grows quadratically with the number of tokens, making standard ViTs computationally expensive. 
To improve training efficiency, DeiT  \cite{b6} introduced data-efficient strategies for ViT training on moderate-scale datasets. Several efficient architectural variants incorporated inductive bias. For example, Swin Transformer \cite{b18} introduces hierarchical shifted window self-attention, PVT \cite{b19} uses a pyramid design to reduce tokens, and CVT introduces convolutional projections to  capture local spatial structures \cite{b20}. Another major research direction focuses on reducing token redundancy during transformer inference. DynamicViT \cite{b2}, A-ViT \cite{b3}, Evo-ViT \cite{b7}, SPViT \cite{b8}, ATS \cite{b9}, IA-RED$^2$ \cite{b10}, Heat-ViT \cite{b14}  show that a substantial portion of tokens can be reduced without severe loss of accuracy. However, directly discarded tokens may still contain important contextual information. A related line of research improves efficiency through token reorganization, merging, pooling, or learned compression. EViT \cite{b4} preserves attentive tokens but fuses all inattentive background tokens into a single representation via an attention-weighted average, losing their spatial arrangement. ToMe \cite{b5}, Joint Token Pruning and Squeezing \cite{b11}, TokenLearner \cite{b13}, and Token Pruning and Pooling \cite{b12},  DeepCompressViT \cite{deepcompress}, and TinyNext-ViT \cite{tinynext} further improve token efficiency through compact token modeling,  while related methods further improve token interaction and fusion for efficient representation learning \cite{b16,b17}. HVCNet \cite{shao2024hybrid} combines local feature extraction with transformers, and HMC \cite{xie2025hybrid} improves the computational efficiency of Vision Transformers.

\begin{figure*}[!t]
    \centering
    \includegraphics[width=0.90\linewidth, height=0.30\textheight,keepaspectratio]{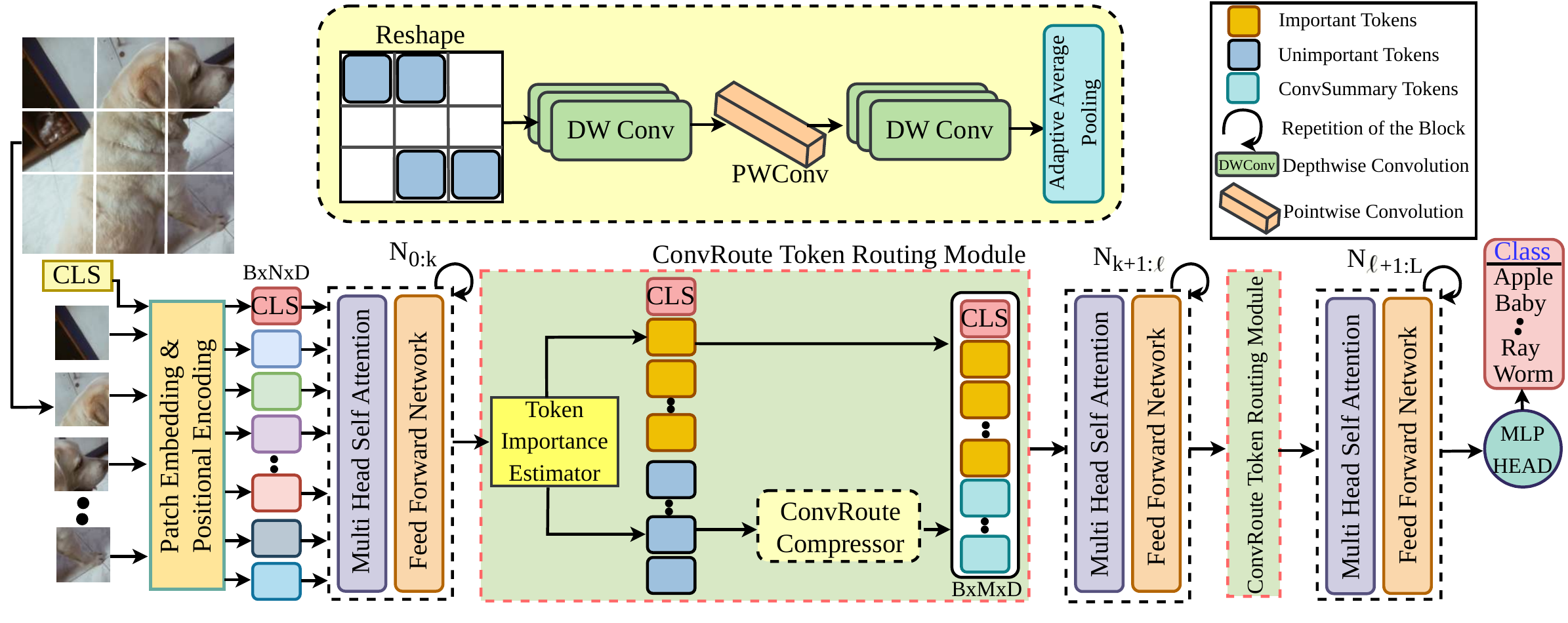}
    
    \caption{Overview of the proposed ConvRoute-ViT architecture. The input image is converted into patch tokens with positional encoding and a CLS token, creating an input size of $B\times N\times D$. After initial transformer processing up to $k^{th}$ transformer layers, the ConvRoute Token Routing Module computes the token importance score and retains the top-R important tokens. The remaining tokens are passed through the ConvRoute Compressor where they are spatially reshaped to generate compact ConvSummary tokens, resulting in $B\times M\times D$  and are processed by transformer layers from $k\!+\!1\!\rightarrow\!l$. A second ConvRoute Token Routing Module is applied at layer $l$ to further reduce the sequence and is then passed through the subsequent transformer blocks for final classification.  
    }
    \label{fig:main}
\end{figure*}

\IEEEpubidadjcol
Motivated by this gap, we propose ConvRoute-ViT, a convolutional token routing framework for efficient vision transformers, as illustrated in Fig. \ref{fig:main}. The proposed mechanism selects informative patch tokens according to CLS-attention scores and retains them for subsequent transformer processing. Rather than discarding, merging, or globally pooling the remaining less informative tokens, ConvRoute recovers their spatial layout and compresses them with a light-weight convolutional module to obtain compact ConvSummary tokens. This design reduces redundancy of tokens while preserving spatially structured contextual information to aid classification. ConvRoute-ViT learns compact, discriminative, and context-preserving token sequences by hierarchically performing the routing process at multiple depths for efficient recognition of images. The routing mechanism can be applied at multiple layers to reduce tokens while maintaining semantically meaningful and contextually important information.

The main contributions of our work:
\begin{itemize}
    \item We introduce the \textbf{ConvRoute token routing module} that estimates the informative patch tokens using CLS-attention scores, allowing the model to focus on task-relevant and discriminative informative regions.
    \item We design the \textbf{ConvRoute Compressor}, which spatially reshapes and compresses them into compact  \textit{ConvSummary} tokens, ensuring useful background structure and complementary context  representations are preserved.
    \item We apply the routing and compression in a \textbf{hierarchical dual-stage manner} at two transformer depths, reducing the token sequence while maintaining rich discriminative representations for final classification.
\end{itemize}

\section{Proposed ConvRoute-ViT Architecture}

\subsection{Preliminaries}
Given an input image $\mathbf{I}\in\mathbb{R}^{H\times W\times C}$, it is divided into non-overlapping patches of size $P\times P$, producing $n=\frac{H}{P}\times\frac{W}{P}.$ Each patch $\mathbf{p}_i\in\mathbb{R}^{P^2C}$ is projected into a $D$-dimensional token as $\mathbf{x}_i=\mathbf{W}_e\mathbf{p}_i$, where $\mathbf{W}_e\in\mathbb{R}^{D\times P^2C}$. A learnable CLS token and positional embeddings form $\mathbf{Z}_0= \mathbf{x}_{cls},\mathbf{x}_1,\dots,\mathbf{x}_n]+\mathbf{E}_{pos},$ where $\mathbf{Z}_0\in\mathbb{R}^{N\times D}$ with $N=n+1$. For the $r$-th transformer block, attention is computed as
\begin{equation}
\mathbf{A}=\text{Softmax}\!\left(\frac{\mathbf{Q}\mathbf{K}^T}{\sqrt{d}}\right),
\end{equation}
where $\mathbf{Q}=\mathbf{Z}_r\mathbf{W}_Q$, $\mathbf{K}=\mathbf{Z}_r\mathbf{W}_K$, and $\mathbf{V}=\mathbf{Z}_r\mathbf{W}_V$. The output is then updated and forwarded to
\begin{equation}
\hat{\mathbf{Z}}_r=\mathbf{Z}_r+\text{MHSA}(\mathbf{Z}_r),\qquad
\mathbf{Z}_{r+1}=\hat{\mathbf{Z}}_r+\text{FFN}(\hat{\mathbf{Z}}_r).
\end{equation}

\definecolor{darkgreen}{RGB}{0,100,0}

\begin{table*}[!t]
\centering
\scriptsize
\caption{Comparison of ConvRoute-ViT under different routing-layer positions $(k,l)$ and keep-ratio settings $(\rho_k,\rho_l)$. Rows denote routing-layer pairs, and each column reports CIFAR-100 and TinyImageNet accuracies for each keep-ratio pair. The base ViT achieves \textcolor{blue}{74.04\%} on CIFAR-100 and \textcolor{darkgreen}{56.80\%} on TinyImageNet. \textcolor{blue}{Blue}: CIFAR-100. \textcolor{darkgreen}{Green}: TinyImageNet.}
\label{tab:my_results}
\renewcommand{\arraystretch}{0.8}
\setlength{\tabcolsep}{2.5pt}
\begin{tabular}{c|cc|cc|cc|cc|cc}
\hline
\multicolumn{1}{c|}{$(k,l)$}
& \multicolumn{2}{c|}{$(\rho_k,\rho_l)=(0.6,0.7)$}
& \multicolumn{2}{c|}{$(\rho_k,\rho_l)=(0.7,0.8)$}
& \multicolumn{2}{c|}{$(\rho_k,\rho_l)=(0.8,0.9)$}
& \multicolumn{2}{c|}{$(\rho_k,\rho_l)=(0.9,0.5)$}
& \multicolumn{2}{c}{$(\rho_k,\rho_l)=(0.9,0.7)$} \\
\hline
(2,4) 
& \textcolor{blue}{73.70} (\textcolor{blue}{-0.34}) & \textcolor{darkgreen}{55.56} (\textcolor{darkgreen}{-1.24})
& \textcolor{blue}{74.06} (\textcolor{blue}{+0.02}) & \textcolor{darkgreen}{55.70} (\textcolor{darkgreen}{-1.10})
& \textcolor{blue}{74.70} (\textcolor{blue}{+0.66}) & \textcolor{darkgreen}{56.62} (\textcolor{darkgreen}{-0.18})
& \textcolor{blue}{74.35} (\textcolor{blue}{+0.31}) & \textcolor{darkgreen}{56.20} (\textcolor{darkgreen}{-0.60})
& \textcolor{blue}{74.07} (\textcolor{blue}{+0.03}) & \textcolor{darkgreen}{56.06} (\textcolor{darkgreen}{-0.74}) \\

(2,5) 
& \textcolor{blue}{73.76} (\textcolor{blue}{-0.28}) & \textcolor{darkgreen}{55.74} (\textcolor{darkgreen}{-1.06})
& \textcolor{blue}{74.18} (\textcolor{blue}{+0.14}) & \textcolor{darkgreen}{56.58} (\textcolor{darkgreen}{-0.22})
& \textcolor{blue}{74.70} (\textcolor{blue}{+0.66}) & \textcolor{darkgreen}{58.04} (\textcolor{darkgreen}{+1.24})
& \textcolor{blue}{74.97} (\textcolor{blue}{+0.93}) & \textcolor{darkgreen}{56.90} (\textcolor{darkgreen}{+0.10})
& \textcolor{blue}{75.27} (\textcolor{blue}{+1.23}) & \textcolor{darkgreen}{57.18} (\textcolor{darkgreen}{+0.38}) \\

(2,6) 
& \textcolor{blue}{73.86} (\textcolor{blue}{-0.18}) & \textcolor{darkgreen}{56.08} (\textcolor{darkgreen}{-0.72})
& \textcolor{blue}{75.54} (\textcolor{blue}{+1.50}) & \textcolor{darkgreen}{56.74} (\textcolor{darkgreen}{-0.06})
& \textcolor{blue}{75.71} (\textcolor{blue}{+1.67}) & \textcolor{darkgreen}{56.74} (\textcolor{darkgreen}{-0.06})
& \textcolor{blue}{74.84} (\textcolor{blue}{+0.80}) & \textcolor{darkgreen}{57.22} (\textcolor{darkgreen}{+0.42})
& \textcolor{blue}{75.03} (\textcolor{blue}{+0.99}) & \textcolor{darkgreen}{57.70} (\textcolor{darkgreen}{+0.90}) \\

(2,7) 
& \textcolor{blue}{74.19} (\textcolor{blue}{+0.15}) & \textcolor{darkgreen}{55.98} (\textcolor{darkgreen}{-0.82})
& \textcolor{blue}{74.21} (\textcolor{blue}{+0.17}) & \textcolor{darkgreen}{56.26} (\textcolor{darkgreen}{-0.54})
& \textcolor{blue}{75.05} (\textcolor{blue}{+1.01}) & \textcolor{darkgreen}{56.96} (\textcolor{darkgreen}{+0.16})
& \textcolor{blue}{75.15} (\textcolor{blue}{+1.11}) & \textcolor{darkgreen}{56.20} (\textcolor{darkgreen}{-0.60})
& \textcolor{blue}{74.93} (\textcolor{blue}{+0.89}) & \textcolor{darkgreen}{57.26} (\textcolor{darkgreen}{+0.46}) \\

(3,5) 
& \textcolor{blue}{74.33} (\textcolor{blue}{+0.29}) & \textcolor{darkgreen}{55.78} (\textcolor{darkgreen}{-1.02})
& \textcolor{blue}{74.54} (\textcolor{blue}{+0.50}) & \textcolor{darkgreen}{56.66} (\textcolor{darkgreen}{-0.14})
& \textcolor{blue}{74.88} (\textcolor{blue}{+0.84}) & \textcolor{darkgreen}{56.58} (\textcolor{darkgreen}{-0.22})
& \textcolor{blue}{74.74} (\textcolor{blue}{+0.70}) & \textcolor{darkgreen}{57.10} (\textcolor{darkgreen}{+0.30})
& \textcolor{blue}{75.42} (\textcolor{blue}{+1.38}) & \textcolor{darkgreen}{57.28} (\textcolor{darkgreen}{+0.48}) \\

(3,6) 
& \textcolor{blue}{74.50} (\textcolor{blue}{+0.46}) & \textcolor{darkgreen}{56.34} (\textcolor{darkgreen}{-0.46})
& \textcolor{blue}{74.90} (\textcolor{blue}{+0.86}) & \textcolor{darkgreen}{56.70} (\textcolor{darkgreen}{-0.10})
& \textcolor{blue}{74.90} (\textcolor{blue}{+0.86}) & \textcolor{darkgreen}{57.28} (\textcolor{darkgreen}{+0.48})
& \textcolor{blue}{75.25} (\textcolor{blue}{+1.21}) & \textcolor{darkgreen}{57.86} (\textcolor{darkgreen}{+1.06})
& \textcolor{blue}{75.15} (\textcolor{blue}{+1.11}) & \textcolor{darkgreen}{57.76} (\textcolor{darkgreen}{+0.96}) \\

(3,7) 
& \textcolor{blue}{74.77} (\textcolor{blue}{+0.73}) & \textcolor{darkgreen}{56.58} (\textcolor{darkgreen}{-0.22})
& \textcolor{blue}{74.93} (\textcolor{blue}{+0.89}) & \textcolor{darkgreen}{56.82} (\textcolor{darkgreen}{+0.02})
& \textcolor{blue}{74.93} (\textcolor{blue}{+0.89}) & \textcolor{darkgreen}{56.86} (\textcolor{darkgreen}{+0.06})
& \textcolor{blue}{75.10} (\textcolor{blue}{+1.06}) & \textcolor{darkgreen}{57.52} (\textcolor{darkgreen}{+0.72})
& \textcolor{blue}{75.41} (\textcolor{blue}{+1.37}) & \textcolor{darkgreen}{57.04} (\textcolor{darkgreen}{+0.24}) \\

(4,6) 
& \textcolor{blue}{74.86} (\textcolor{blue}{+0.82}) & \textcolor{darkgreen}{56.16} (\textcolor{darkgreen}{-0.64})
& \textcolor{blue}{74.65} (\textcolor{blue}{+0.61}) & \textcolor{darkgreen}{56.78} (\textcolor{darkgreen}{-0.02})
& \textcolor{blue}{75.39} (\textcolor{blue}{+1.35}) & \textcolor{darkgreen}{57.18} (\textcolor{darkgreen}{+0.38})
& \textcolor{blue}{75.00} (\textcolor{blue}{+0.96}) & \textcolor{darkgreen}{56.88} (\textcolor{darkgreen}{+0.08})
& \textcolor{blue}{75.07} (\textcolor{blue}{+1.03}) & \textcolor{darkgreen}{57.70} (\textcolor{darkgreen}{+0.90}) \\

(4,7) 
& \textcolor{blue}{74.75} (\textcolor{blue}{+0.71}) & \textcolor{darkgreen}{57.20} (\textcolor{darkgreen}{+0.40})
& \textcolor{blue}{74.85} (\textcolor{blue}{+0.81}) & \textcolor{darkgreen}{57.42} (\textcolor{darkgreen}{+0.62})
& \textcolor{blue}{74.55} (\textcolor{blue}{+0.51}) & \textcolor{darkgreen}{57.28} (\textcolor{darkgreen}{+0.48})
& \textcolor{blue}{74.89} (\textcolor{blue}{+0.85}) & \textcolor{darkgreen}{57.56} (\textcolor{darkgreen}{+0.76})
& \textcolor{blue}{74.82} (\textcolor{blue}{+0.78}) & \textcolor{darkgreen}{57.02} (\textcolor{darkgreen}{+0.22}) \\
\hline
\end{tabular}
\end{table*}

\subsection{ConvRoute Token Routing Module}

For a transformer with $L$ layers, ConvRoute routing is applied at two layers $k$ and $l$, where $0<k<l<L$. The $k$ layers process the full token sequence before reduction. After layer $k$, the token sequence is represented as $\mathbf{Z}^{(k)}=[\mathbf{z}{cls}^{(k)},\mathbf{X}^{(k)}]$, where $\mathbf{z}{cls}^{(k)}\in\mathbb{R}^{D}$ is the CLS token and $\mathbf{X}^{(k)}\in\mathbb{R}^{(N-1)\times D}$ are the patch tokens. The importance score of the $i$-th patch token is computed from CLS-attention as
\begin{equation}
s_i^{(k)}=\frac{1}{h_A}\sum_{h=1}^{h_A}A^{(k,h)}_{cls,i},
\end{equation}
where $h_A$ is the number of attention heads and $A^{(k,h)}_{cls,i}$ denotes the attention from the CLS token to the $i$-th patch token in the $h$-th head. Based on the keep ratio $\rho_k$, the top $R_k=\lfloor\rho_k(N-1)\rfloor$ tokens are retained as $\mathbf{T}_{imp}^{(k)}\in\mathbb{R}^{R_k\times D}$, while the remaining tokens are denoted by $\mathbf{T}_{low}^{(k)}$. Instead of discarding $\mathbf{T}_{low}^{(k)}$, ConvRoute restores these tokens to their original patch-grid positions to form $\mathbf{T}_{map}^{(k)}\in\mathbb{R}^{D\times H_p\times W_p}$, where $H_p=\frac{H}{P}$ and $W_p=\frac{W}{P}$. The restored map is compressed through the ConvRoute Compressor as
\begin{equation}
\mathbf{F}^{(k)}=
\text{DWConv}_{3\times3}\Big(
\text{PWConv}_{1\times1}\big(
\text{DWConv}_{3\times3}(\mathbf{T}_{map}^{(k)})
\big)\Big),
\end{equation}
\begin{equation}
\mathbf{T}_{conv}^{(k)}=
\text{Flatten}\big(\text{AdaptiveAvgPool}_{p_h\times p_w}(\mathbf{F}^{(k)})\big),
\end{equation}
where $\mathbf{T}_{conv}^{(k)}\in\mathbb{R}^{S_{k}\times D}$ and $S_{k}=p_h\times p_w$ is the number of ConvSummary tokens. The depthwise convolutions preserve local spatial neighborhoods, while the pointwise convolution captures channel interactions, enabling low-importance tokens to be summarized without losing contextual information.

The routed sequence after the first stage is
\begin{equation}
\tilde{\mathbf{Z}}^{(k)}=[\mathbf{z}_{cls}^{(k)},\mathbf{T}_{imp}^{(k)},\mathbf{T}_{conv}^{(k)}]\in\mathbb{R}^{M_k\times D},    
\end{equation}

where $M_k=[1 + R_k + S_k]$ and $M_k\leq N$. The same routing process is applied again at layer $l$, producing
$
\tilde{\mathbf{Z}}^{(l)}=[\mathbf{z}_{cls}^{(l)},\mathbf{T}_{imp}^{(l)},\mathbf{T}_{conv}^{(l)}]\in\mathbb{R}^{M_l\times D},
$
with $M_l\leq M_k\leq N$. Thus, ConvRoute progressively reduces token redundancy while preserving discriminative tokens and compact spatial-context information for final classification.

\section{Experimental Settings and Results} \label{sec:exp_settings}

\subsection{Datasets and Implementation Details} 
\noindent \textbf{Datasets}: We conducted our experiments on CIFAR-100 \cite{b21} contains 60,000 color images of size  $32 \times 32$ from 100 distinct classes, with 50,000 training and 10,000 test images and TinyImageNet \cite{b22} contains 200 classes, each with 500 training images and 50 validation images of size $64 \times 64$. 
\noindent \textbf{Experimental Settings}: Models are trained from scratch on an NVIDIA GeForce RTX 4090 GPU for 100 epochs using AdamW with an learning rate of 0.003, weight decay of $5\times10^{-2}$, cosine annealing, a batch size of 128 with 12 attention heads.

\begin{figure}
    \centering
    \includegraphics[width=0.98\linewidth]{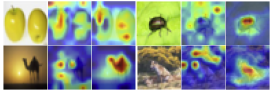}
 
    \caption{Attention heatmaps for representative CIFAR-100 samples \cite{b21}. Each example consists of three panels arranged from left to right: the input image, the base ViT attention map, and the ConvRoute-ViT attention map.}
    \label{fig:amp}
\end{figure}

\subsection{Experimental Results}

Table \ref{tab:my_results} shows that ConvRoute-ViT consistently improves the base ViT across most routing-layer pairs. The best CIFAR-100 accuracy is obtained at layers $(2,6)$ with $(\rho_k,\rho_l)=(0.8,0.9)$, achieving 75.71\% (+1.67), while the best TinyImageNet result is obtained at $(2,5)$ with $(\rho_k,\rho_l)=(0.8,0.9)$, reaching 58.04\% (+1.24). Improved gains are also observed at intermediate routing positions such as $(3,5)$ and $(3,6)$, where the token features are sufficiently discriminative to separate important object regions from less informative context, allowing ConvRoute to remove redundancy without losing semantic information. While routing in early layers, such as at $(2,4)$, removes useful low-level details before stable feature formation, which explains the weaker gains, since low-level visual details are reduced before the formation of feature representations. Overall, the results indicate that token reduction depends on selecting routing layers that reduce less informative tokens while preserving useful contextual information. 


\subsection{Attention Visualization Analaysis}
Figure \ref{fig:amp} presents four representative CIFAR100 \cite{b21} examples comparing the attention map visualization  of the base ViT and our proposed ConvRoute-ViT. In the first example, the base ViT wrongly predicts \textit{apple} as \textit{orange} by focusing on the smooth, round color patterns, whereas ConvRoute-ViT attends to highly discriminative regions and predicts \textit{apple}. In the second example,  the base ViT scatters its attention to the background areas, leading to \textit{caterpillar} instead of \textit{beetle,} while our model concentrates over the body and preserves the fine-grained structure required for correct classification. In the third example, the base ViT predicts \textit{orange} for \textit{camel}, by concentrating on the bright round background regions, while ConvRoute-ViT places a coherent high response over the camel and correctly predicts \textit{camel}.  For the fourth example, the base ViT focuses on textured background regions and doenot focus on the main body of the fox, whereas ConvRoute-ViT produces a compact and coherent response over the fox region and gives the correct class. Overall, these visualizations show that ConvRoute-ViT improves classification by suppressing misleading background tokens and emphasizing the most informative object-centric regions.

\begin{figure}
    \centering
    \includegraphics[width=0.98\linewidth]{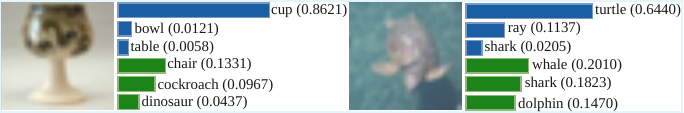}
  
    \caption{Comparison of prediction confidence between the proposed model and the base ViT on CIFAR-100. Input images are shown on the left, and the top-3 predictions on the right (\textcolor{deepblue}{blue}: proposed; \textcolor{deepgreen}{green}: base ViT). }
    \label{fig:confidence}
\end{figure}

\begin{table*}[!t]
\centering
\scriptsize
\setlength{\tabcolsep}{0.95pt}
\renewcommand{\arraystretch}{1.2}
\caption{Effect of keep-ratio pair settings $(\rho_k,\rho_l)$ at the 4th and 6th routing layers. Columns shows different pairs, and rows report CIFAR-100, TinyImageNet, and GFLOPs. The baseline achieves \textcolor{blue}{74.04\%} on CIFAR-100, \textcolor{darkgreen}{56.80\%} on TinyImageNet, and 0.1890 GFLOPs.}

\begin{tabular}{c|cccccccccccccccccc}
\hline
$(\rho_k,\rho_l)$ 
& (0.4,0.5) & (0.5,0.4) & (0.5,0.5)
& (0.5,0.6) & (0.6,0.5) & (0.6,0.6)
& (0.6,0.7) & (0.7,0.6) & (0.7,0.7)
& (0.7,0.8) & (0.8,0.7) & (0.8,0.8)
& (0.8,0.9) & (0.9,0.5) & (0.9,0.6) & (0.9,0.7)
& (0.9,0.8) & (0.9,0.9) \\
\hline
\textbf{\textcolor{blue}{CIFAR-100}}    
& \textcolor{blue}{74.28} & \textcolor{blue}{74.34} & \textcolor{blue}{74.55}
& \textcolor{blue}{74.25} & \textcolor{blue}{74.70} & \textcolor{blue}{74.55}
& \textcolor{blue}{74.86} & \textcolor{blue}{74.75} & \textcolor{blue}{74.87}
& \textcolor{blue}{74.65} & \textcolor{blue}{75.05} & \textcolor{blue}{74.40}
& \textcolor{blue}{75.39} & \textcolor{blue}{75.00} & \textcolor{blue}{75.01} & \textcolor{blue}{75.07}
& \textcolor{blue}{74.01} & \textcolor{blue}{75.06} \\
\textbf{\textcolor{darkgreen}{TinyImageNet}} 
& \textcolor{darkgreen}{55.48} & \textcolor{darkgreen}{56.36} & \textcolor{darkgreen}{55.34}
& \textcolor{darkgreen}{56.56} & \textcolor{darkgreen}{56.82} & \textcolor{darkgreen}{56.96}
& \textcolor{darkgreen}{56.16} & \textcolor{darkgreen}{56.90} & \textcolor{darkgreen}{57.12}
& \textcolor{darkgreen}{56.78} & \textcolor{darkgreen}{56.32} & \textcolor{darkgreen}{57.92}
& \textcolor{darkgreen}{57.18} & \textcolor{darkgreen}{56.88} & \textcolor{darkgreen}{57.20} & \textcolor{darkgreen}{57.70}
& \textcolor{darkgreen}{57.22} & \textcolor{darkgreen}{57.26} \\
\textbf{GFLOPs}       
& 0.1239 & 0.1287 & 0.1325
& 0.1353 & 0.1395 & 0.1433
& 0.1472 & 0.1504 & 0.1553
& 0.1602 & 0.1653 & 0.1712
& 0.1763 & 0.1797 & 0.1697 & 0.1618
& 0.1736 & 0.1858 \\
\hline
\end{tabular}
\label{tab:kr}
\end{table*}

\subsection{Effect of Keep Ratio on Accuracy and Complexity}
Table \ref{tab:kr}  shows a trade-off between classification performance and computational cost. The low complexity settings $ \rho = (0.4, 0.5)$ reduce the GFLOPs to 0.1239, corresponding to a 34.44\% reduction over the base model, while improving CIFAR-100 accuracy from 74.04\% to 74.28\%. As the keep ratio increases, the model generally preserves more informative tokens, which improves accuracy. On CIFAR-100, the best result is obtained at $ \rho = (0.8, 0.9)$ with 75.39\% (+1.35) with 6.72\% reduction, while on TinyImageNet the best performance is achieved at $ \rho = (0.8, 0.8)$ with 57.92\% (+1.12), corresponding to a 9.42\% reduction. Overall, moderate keep-ratio settings provide the most favorable balance between efficiency and accuracy, as they retain a sufficient number of discriminative tokens while still reducing a substantial amount of computation from less informative regions.

\begin{table}[!t]
\centering
\scriptsize
\setlength{\tabcolsep}{2pt}
\renewcommand{\arraystretch}{0.95}
\caption{Comparison with representative CNN and Transformer baselines on CIFAR-100 and TinyImageNet. Results are taken from \cite{hsvit}, and models marked with {\dag} are implemented from scratch.}
\label{tab:comparison}
\begin{tabular}{lccc}
\hline
\textbf{Model} & \textbf{GFLOPs} & \textbf{CIFAR-100} & \textbf{TinyImageNet} \\
\hline
ResNet-50 \cite{he2016deep}                & 0.7 & 71.03 & 56.02 \\
EfficientNet-B3 \cite{tan2019efficientnet} & 0.2 & 68.77 & 48.85 \\
MobileNetV2-1.4 \cite{mobilenetv2}         & 0.1 & 69.92 & 54.12 \\
CvT-21 \cite{cvt}                          & 1.7 & 63.32 & 49.97 \\
HSViT-C3A4 \cite{hsvit}                    & 1.3 & 72.46 & 56.73 \\
CaiT {\dag} \cite{touvron2021going}        & 0.6 & 72.44 & 57.59 \\
PVT-S {\dag} \cite{wang2021pyramid}        & 0.5 & 61.52 & 49.26 \\
ToMe {\dag} \cite{24}                      & 0.4 & 60.25 & 44.20 \\
EViT {\dag} \cite{b4}                      & 0.2 & 65.94 & 52.24 \\
DeepCompress-ViT {\dag} \cite{deepcompress}       & --  & 58.39 & 65.02 \\
TinyNeXt {\dag} \cite{tinynext}                   & --  & 49.45 & 43.34 \\
\hline
Ours ConvRoute-ViT                         & 0.1 & 75.42 & 58.56 \\
\hline
\end{tabular}
\end{table}

\subsection{Confidence Analysis}
Figure \ref{fig:confidence} compares the top-3 confidence of the predicted classes of the base ViT and our proposed ConvRoute-ViT across two distinct classes from the CIFAR100 dataset \cite{b21}. For the first example \textit{cup}, ConvRoute-ViT assigns the highest confidence correctly to \textit{the cup} with a score of 0.8621, whereas the base ViT spreads its confidence over the incorrect classes such as \textit{chair} (0.1331), \textit{cockroach} (0.0967), and \textit{dinosaur} (0.0437). The second example, ConvRoute-ViT, correctly predicts \textit{turtle} (0.6440), whereas the baseline favors incorrect marine categories, assigning scores to \textit{whale} (0.2010), \textit{shark} (0.1823), and \textit{dolphin} (0.1470). Thus, the confidence scores show that ConvRoute-ViT produces more accurate and better concentrated class probabilities by preserving informative object-centric tokens and reducing the effect of misleading background or semantically irrelevant tokens.

\subsection{Analysis of the Token Reduction}
\begin{figure}
    \centering
    \includegraphics[width=0.98\linewidth]{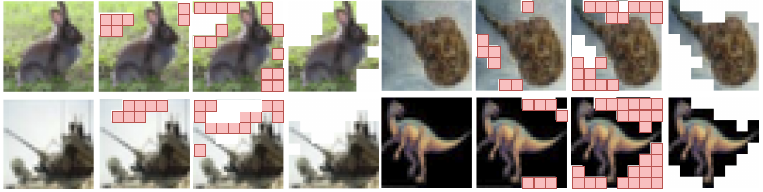}
  
    \caption{Four representative CIFAR-100 examples illustrating the proposed ConvRoute mechanism. From left to right: input image, tokens removed at layer $k$, additional tokens removed at layer $l$, and the final compact token representation, which is further processed and passed for final classification.}
    \label{fig:tokens}
\end{figure}

Figure \ref{fig:tokens} illustrates how the proposed ConvRoute mechanism progressively reduces the token sequence across multiple routing stages. For this visualization, we set keep rates to $\rho=0.9$ at the $3^{rd}$ layer and $\rho=0.7$ at the $5^{th}$ layer. In each row, the first panel shows the original input image. The second panel highlights the less informative tokens routed out at the $3^{rd}$ layer. The third panel shows the further tokens, which are reduced at the $5^{th}$ layer, with both sets of removed tokens depicted as red patches. The final image presents the cumulative reduction of patch tokens after the routing layers. These visualizations show that ConvRoute-ViT reduces tokens that are  mainly concentrated in less informative background regions, while the visually important object regions are significantly preserved. Such token reduction enables the model to reduce redundancy  while retaining the most discriminative and important regions for efficient transformer processing.

\subsection{Results Comparison }

The results in Table \ref{tab:comparison} demonstrate that ConvRoute-ViT achieves the best overall performance among the compared CNN and Transformer baselines while maintaining only 0.1 GFLOPs. It achieves 75.42\% accuracy on CIFAR-100 and 58.56\% on TinyImageNet, outperforming all competing methods by margins ranging from +2.96\% to +25.97\% on CIFAR-100 and +0.17\% to +15.22\% on TinyImageNet. In particular, ConvRoute-ViT surpasses recent efficient Transformer models such as CaiT \cite{touvron2021going}, PVT-S \cite{wang2021pyramid}, ToMe \cite{24}, DeepCompress-ViT \cite{deepcompress}, and TinyNeXt \cite{tinynext} while requiring only 0.1 GFLOPs. These results demonstrate that ConvRoute-ViT effectively balances computational efficiency and recognition performance by reducing redundant tokens while preserving informative contextual representations. To further evaluate deployment efficiency, we evaluated the best ConvRoute-ViT configuration on CIFAR-100 against the corresponding base ViT on a single RTX 4090 GPU. ConvRoute-ViT reduced the total inference time from 0.7653\,s to 0.4877\,s, decreased the latency from 0.077 to 0.049\,ms/image, increased the throughput from 13,066.55 to 20,506.16 images/s, and lowered the peak GPU memory usage from 399.09\,MB to 357.22\,MB improving both classification accuracy and deployment efficiency.

\section{Conclusion}
In this paper, we presented ConvRoute-ViT, an efficient vision transformer framework that addresses the redundancy of uniform token processing through a convolution token routing module. By separating the token sequence into informative and compact convolutional summary tokens, our proposed mechanism reduces the sequence length in transformer layers without completely dropping the residual context information in the less informative token patches. The overall findings show that effective token reduction should not only depend on removing tokens but should also preserve the supporting visual features that remain useful  for classification. The quantitative results, along with the attention, confidence, and token reduction analyses, demonstrate that our ConvRoute-ViT learns more focused and compact representations. Overall, our proposed framework demonstrates that convolution-guided token summarization provides an effective direction for efficient vision transformer design.

Our ConvRoute-ViT preserves the discriminative CLS-attended tokens and compresses the less informative tokens into the condensed ConvSummary representations, which can improve the recognition performance by reducing the token redundancy without discarding useful contextual information. This hierarchical routing strategy enables the deeper transformer layers to process shorter but semantically richer sequences of tokens, leading to higher accuracy and computational efficiency. However, the evaluation is currently performed only on CIFAR-100 and TinyImageNet. In future work we plan to study scalability of ConvRoute-ViT on larger datasets (e.g. ImageNet-1K); conduct statistical analysis over multiple runs; perform extensive ablation studies on ConvSummary tokens; explore deployment on resource-constrained edge devices; and extend the framework to video transformers for efficient spatio-temporal token routing.


\bibliographystyle{IEEEtran}
\bibliography{ref}

\end{document}